\documentclass[11pt,a4paper]{article}
\usepackage[hyperref]{emnlp-ijcnlp-2019}
\usepackage{latexsym}
\usepackage{times}
\usepackage{soul}
\usepackage{url}
\usepackage[utf8]{inputenc}
\usepackage{caption}
\usepackage{graphicx}
\usepackage{subfigure}
\usepackage{amsmath}
\usepackage{booktabs}
\usepackage{natbib}
\usepackage{xcolor}
\usepackage{fancyhdr,graphicx,amssymb}
\usepackage[ruled,vlined]{algorithm2e}
\usepackage{multirow}
\usepackage{url}
\usepackage{color}

\usepackage{tikz}
\usepackage{subfigure}
\usepackage{xcolor}
\usepackage{tcolorbox}

\usepackage{helvet}  
\usepackage{courier}  

\usetikzlibrary{shadows}
\usetikzlibrary{shadows.blur}

\aclfinalcopy

\title{Transfer Learning Enhanced Single-choice Decision for Multi-choice Question Answering}

\author{
	Chenhao Cui$^{1}$
Yufan Jiang$^{2}$,
  Shuangzhi Wu$^{2}$,
  Zhoujun Li$^{3}$,\\
  $^{1}$School of Cyber Science and Technology, Beihang University\\ 
  $^{2}$Tencent Cloud Xiaowei\\ 
  $^{3}$School of Computer Science and Engineering, Beihang University \\
  {\tt
  	cuich@buaa.edu.cn
  }\\
  {\tt
        \{frostwu,garyyfjiang\}@tencent.com 
  }\\
  {\tt
       lizj@buaa.edu.cn
  }\\
}

\date{}

\begin{document}
\maketitle

\begin{abstract}
Multi-choice Machine Reading Comprehension (MMRC) aims to select the correct answer from a set of options based on a given passage and question. The existing methods employ the pre-trained language model as the encoder, share and transfer knowledge through fine-tuning.
These methods mainly focus on the design of exquisite mechanisms to effectively capture the relationships among the triplet of passage, question and answers. It is non-trivial but ignored to transfer knowledge from other MRC tasks such as SQuAD due to task specific of MMRC.
In this paper, we reconstruct multi-choice to single-choice by training a binary classification to distinguish whether a certain answer is correct. Then select the option with the highest confidence score as the final answer. Our proposed method gets rid of the multi-choice framework and can leverage resources of other tasks. We construct our model based on the ALBERT-xxlarge model and evaluate it on the RACE and DREAM datasets. Experimental results show that our model performs better than multi-choice methods. In addition, by transferring knowledge from other kinds of MRC tasks, our model achieves state-of-the-art results in both single and ensemble settings.

\end{abstract}

\section{Introduction}

Machine reading comprehension (MRC) has been a heated task that involves training a model to comprehend text and answer questions written in natural languages. In the last several years, various datasets and methods have been proposed. 
Given passages and questions, MRC can be divided into several types according to the form of answers.
Extractive tasks require the model to find the correct span from the context as the answer, such as SQuAD1.1, SQuAD2.0 and CoQA \cite{rajpurkar-etal-2016-squad,rajpurkar-etal-2018-know,reddy-etal-2019-coqa}.
Different from the above, multi-choice MRC (MMRC) is to select the right option according to the context and question. RACE is an MMRC dataset \cite{lai-etal-2017-race}, which is extracted from middle and high school English examinations in China. Table. \ref{tab:example} shows an example passage and related two questions from RACE. The key difference between RACE and previously released machine comprehension datasets is that the answers in RACE often cannot be directly extracted from the passages. Thus, answering these questions needs inferences. Different task types require different decoders. 
Extractive MRC requires the model to predict the start and end position from the context as the answer, such as SQuAD. 
However, MMRC models apply classifiers to select options. 
In addition, the number of options in the MMRC dataset varies, leading to differences in the model structure. Models trained on a 4-option dataset often face challenges when directly applied to a 3-option dataset, and they require retraining.
Due to differences in task types and the number of options, models trained on a specific multiple-choice dataset often cannot transfer knowledge from other existing datasets.

\begin{table}[t!]
\centering
  \caption{An example passage and two related multi-choice questions. The ground-truth answers are in \textbf{bold}.}
  \begin{center}
  \label{tab:example}
  \begin{tabular}{p{7.2cm}}
  \toprule
  \textbf{Passage}: For the past two years, 8-year-old Harli Jordean from Stoke Newington, London, has been selling marbles . His successful marble company, Marble King, sells all things marble-related - from affordable tubs of the glass playthings to significantly expensive items like Duke of York solitaire tables - sourced, purchased and processed by the mini-CEO himself. "I like having my own company. I like being the boss," Harli told the Mirror....Tina told The Daily Mail. "At the moment he is annoying me by creating his own Marble King marbles - so that could well be the next step for him." \\
  \midrule
  \textbf{Q1}: Harli's Marble Company became popular as soon as he launched it because \_\_\_ .     \\
  \textbf{A}: It was run by "the world's youngest CEO"      \\
  \textbf{B: It filled the gap of online marble trade}    \\
  \textbf{C}: Harli was fascinated with marble collection   \\
  \textbf{D}: Harli met the growing demand of the customers \\
   \midrule
  \textbf{Q2}: How many mass media are mentioned in the passage?    \\
  \textbf{A}: One      \\
  \textbf{B}: Two     \\
  \textbf{C: Three}   \\
  \textbf{D}: Four \\
  \bottomrule
  \end{tabular}
  \end{center}
\end{table}

In recent years, pre-trained language models such as BERT \cite{devlin-etal-2019-bert}, RoBERTa \cite{liu2019roberta}, ALBERT \cite{Lan2020ALBERT} have achieved great success on MMRC tasks. Notably, Megatron-LM \cite{shoeybi2019megatron} which is a 48 layers BERT with 3.9 billion parameters yields the highest score on the RACE leaderboard in both single and ensemble settings. The key point to model MMRC is: first encode the context, question and options with a language model like BERT, then add a matching network on top of BERT to score the options. Generally, the matching network can be various. Ran \textit{et. al.}~\cite{ran2019option} proposes an option comparison network (OCN) to compare options at the word-level to better identify their correlations to help reason. Zhang \textit{et. al.}~\cite{Zhang_Zhao_Wu_Zhang_Zhou_Zhou_2020} proposes a dual co-matching network (DCMN) that models the relationship among passage, question and answer options bidirectionally. All these matching networks show promising improvements based on pre-trained language models. One point they have in common is that the answer together with the distractors are jointly considered which we name multi-choice models. We argue that the options can be considered separately for two reasons, 1) when humans work on MMRC problems, they always consider the options one by one and select the one with the highest confidence. 2) MMRC suffers from the data scarcity problem. Multi-choice models are inconvenient to take advantage of other MRC datasets.

In this paper, we propose a single-choice model for MMRC. Our model considers the options separately. The key component of our method is a binary classification network on top of pre-trained language models. For each option of a given context and question, we calculate a confidence score. Then we select the one with the highest score as the final answer. In both training and decoding, the right answer and the distractors are modeled independently. Our proposed method gets rid of the multi-choice framework, and can leverage the amount of other resources. Taking SQuAD as an example, we can take a context, one of its questions and the corresponding answer as a positive instance for our classification with golden label 1. In this way, many QA datasets can be used to enhance RACE. Experimental results show that single-choice model performs better than multi-choice models, in addition by transferring knowledge from other QA datasets, our single model achieves 90.7\% and ensemble model achieves 91.4\%, both are the best score on the leaderboard. In addition, we also achieve the best results on the DREAM dataset.

\section{Related work}

\subsection{Machine reading comprehension}

The types of questions and answers in machine reading comprehension are flexible and diverse, such as arithmetic, abstract, commonsense, logical reasoning\cite{DBLP:journals/fcsc/SunLDNC22,DBLP:conf/emnlp/GhosalMMP22}, language inference, and sentiment analysis\cite{DBLP:journals/fcsc/ZengLCM23}. Generally, MRC can be divided into several types according to the form of answers. Cloze-style requires the model to decide which word or entity is the most appropriate for the placeholder in the question, including CNN/DM \cite{NIPS2015_afdec700}, CBT \cite{hill2016goldilocks}.  Multi-choice requires the model to select the correct one from multiple assumptions according to the context, such as MCTest \cite{richardson2013mctest}, RACE \cite{lai-etal-2017-race}, DREAM \cite{sun-etal-2019-dream}. 
The extractive category requires the model to find the correct span from the context as the answer, such as
SQuAD \cite{rajpurkar-etal-2016-squad}, SQuAD2.0\cite{rajpurkar-etal-2018-know}.The answer of Free-Form is human-generated, be any free-text form, such as NarrativeQA \cite{kovcisky2018narrativeqa}, Dureader \cite{he2018dureader}.

\subsection{Deep learning-based approaches}

Attention mechanism was first proposed by \cite{DBLP:journals/corr/BahdanauCB14} and applied to machine translation. Since then, attention mechanism has been introduced to various tasks of natural language processing. Before the large pre-trained models (PrLMs) are proposed, methods are based on traditional encoders such as s Long Short-Term Memory (LSTM) \cite{LSTM} to model representations. After transformer \cite{NIPS2017_3f5ee243} is proposed and proved to be very effective, many PrLMs using different training strategies have been trained to constantly refresh state-of-the-art of NLP tasks. Based on PrLMs, numerous methods have been proposed to capture and infer the relationships among passages,
questions, and the corresponding candidate options. 
Ran \textit{et. al.}~\cite{ran2019option} distinguishes answer options by modeling relationship and interaction among options.
Kim \textit{et. al.}~\cite{kim2020learning} improves performance by training the model to eliminate wrong options.
Zhang \textit{et. al.}~\cite{Zhang_Zhao_Wu_Zhang_Zhou_Zhou_2020} propose a dual co-matching network, using
the key sentence selection mechanism to select more important sentences for
answering questions from the passage, considering comparison information among answers to improve the matching representations. 
Jin \textit{et. al.}~\cite{Jin_Gao_Kao_Chung_Hakkani-tur_2020} propose a multi-stage multi-task learning framework. Coarse-tuning stage relies on two out-of-domain datasets, multi-task learning stage relies on a larger in-domain dataset to enable the model to achieve better generalization with a limited amount of data. 
Zhu \textit{et. al.}~\cite{zhu2021duma} propose a DUal Multi-head Co-Attention model, considering each other’s focus from the standpoint of passage and question-option pairs by calculating the attention score between them.
Some works introduce a unified framework to explore transfer learning. By converting tasks into text-to-text format, the model can handle multiple types of tasks \cite{2020unifiedqa, 2020t5}. 

\subsection{Pre-trained language models}

All models try to mine more information from the context to improve the model performance. However, it must be noted that human beings often introduce knowledge outside the context when considering problems. For example, when referring to the White House, we usually think of the President of the United States, while the Great Wall represents China. Transfer learning is a way to solve this problem.  It has been widely used and proved to be effective in the NLP. The most common example of transfer learning is the use of large pre-trained models.

BERT \cite{devlin-etal-2019-bert} applies a Transformer encoder to attend to bi-directional contexts during pre-training. BERT proposes a masked language modeling (MLM) objective and  a next-sentence-prediction (NSP) objective to assist in training.
RoBERTa \cite{liu2019roberta} makes a few changes to the released BERT model and achieves substantial improvements, including removing the NSP objective and dynamically changing the masked positions during pretraining.
ALBERT \cite{Lan2020ALBERT}  proposes two parameter-reduction techniques (factorized embedding parameterization and cross-layer parameter sharing) to lower memory consumption and
speed up training. Besides, ALBERT instead uses a sentence-order prediction
(SOP) objective.

\section{Methods}

\begin{figure*}[htp]
    \centering
    \subfigure[ Multi-choice Model]{
        \includegraphics[width=0.9\textwidth]{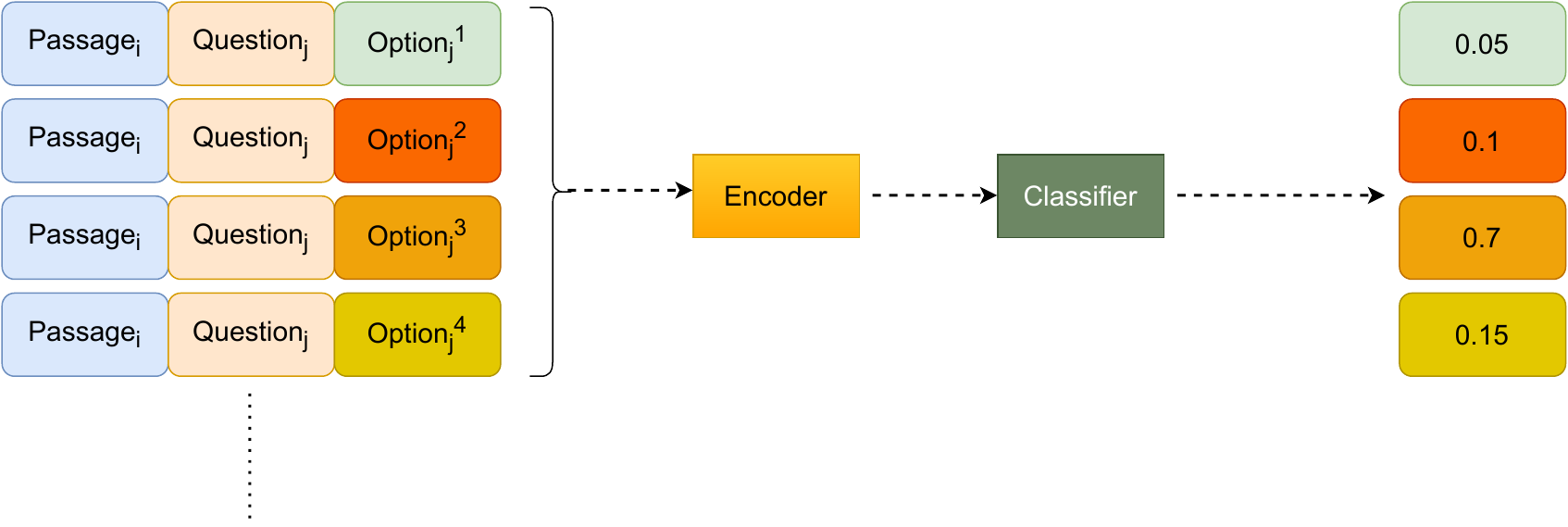}
        \label{mc_model}
    }
    \subfigure[Single-choice Model]{
    \includegraphics[width=0.9\textwidth]{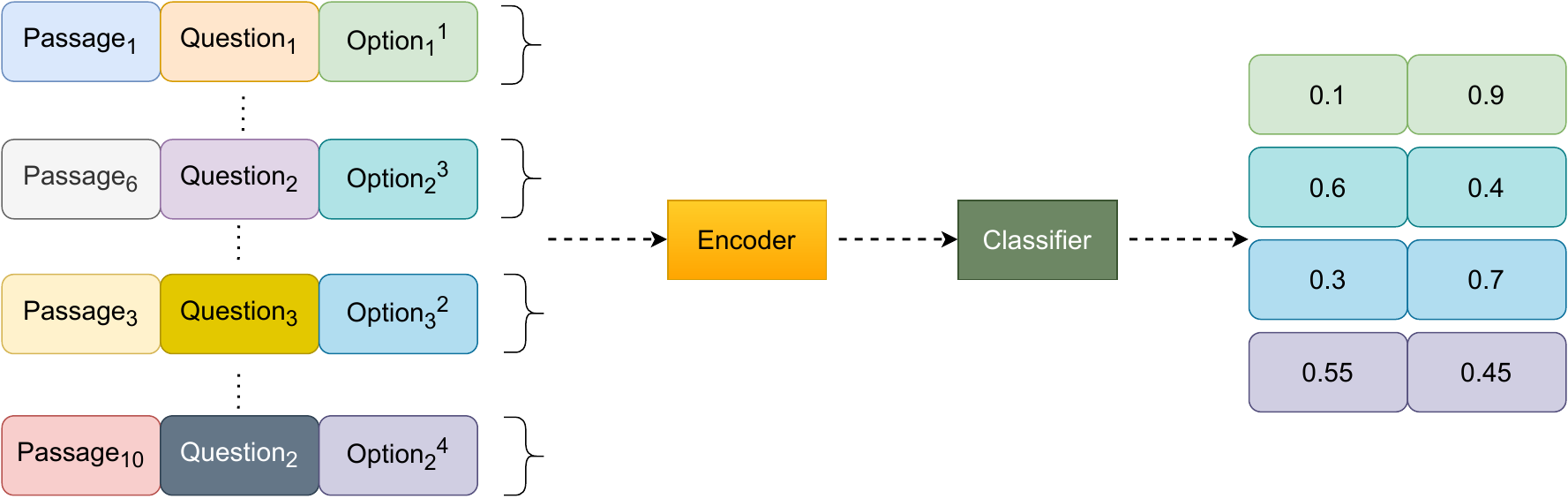}
        \label{sc_model}
    }
    \caption{An overview of standard Multi-choice Model and Our Single-choice Model.}
    \label{fig.1}
\end{figure*}

\subsection{Task Description}

Multi-choice MRC can be represented as a triple $<P, Q, A>$, where $P$ is an article consisting of multiple sentences, $Q$ is a question asked upon the article and $A=\{A_1,...A_n\}$ is the option set with $n$ answer candidates. Only one answer in $A$ is correct and others are distractors. The purpose of the MMRC is to select the right one.

\subsection{Multi-choice Model}

Previous works have verified the effectiveness of Pre-trained language models such as BERT, XLNet, RoBERTa, and ALBERT in Multi-choice MRC tasks. Pre-trained language models are used as an encoder to get the global context representation. After that, a decoder is employed to find the correct answer given all the information contained in the global representation.

First, the passage $P$, question $Q$ and one of the answer options $A_i$ are concatenated into a sequence of length $l$, defined as $\{P \oplus Q \oplus A_i\}$. For a question with $n$ answer
options, the complete training sample is $n$ token sequences of length $l$. Afterward, it will be fed into the encoder, and the $[CLS]$ vector of the last layer is usually used as the context representation $H = \{H_{CLS1}, ..., H_{CLSn}\} \in \mathbb R^{n\times d} $, which is used to classify which is the correct answer given passage and question(see Figure \ref{mc_model}). A top-level
classifier $C$ as the decoder is added to compute the probability $p(\{A_1,...A_n\}|P, Q)$ for all the answers and 
can be computed as follows:
\begin{equation}
p(\{A_1,...A_n\}|P,Q) = C (WH + b) \\
\end{equation}
where $W$ is the weight and $b$ is the bias. 
The option with the highest probability will be considered as the answer to the question.
Parameter matrices are finetuned based on pre-trained language model. Let the ground truth $y$ denote the index of correct answer in candidates, the cross entropy loss function is formulated as:
\begin{eqnarray}
\text{loss} &=& -\sum y \log (p)
\end{eqnarray}

\subsection{Single-choice Model}
For multi-choice models, all input sequences with the same passage and question are tied together, each training sample contains much duplicate content, which may degrade the diversity in each training step. For example, the passage with multiple sentences repeats $n$ times in a single training sample, only the options are different.
Besides, Multi-choice models are sensitive to the relative difficulty between the options because models only need minimally distinguish the correct option from the incorrect options. It means that the models do not necessarily think that the correct option is correct and the wrong options are wrong, but only need to give the correct one a higher score than the wrongs.
Moreover, multi-choice methods need to fix the data format that each question must have the same number of options which is also inconvenient to take advantage of other MRC datasets.

Alternatively, we reconstruct the multi-choice to single-choice. The single-choice model
 is trained to predict a high score for the correct option triplet $\{P \oplus Q \oplus A_g\}$, and low for the other $n - 1$ triplets. The model needs to distinguish whether the answer is correct separately without considering other options in the candidate set. In this way, we keep the diversity in training batches and relax the constraints on multi-choice framework.

Different from the multi-choice models, the complete training sample for single-choice model is just the concatenation $\{P \oplus Q \oplus A_i\}$. Afterward, each sample sequence will be encoded, as shown in Figure \ref{sc_model}. Instead of using the $[CLS]$ vector of the last layer, we propose layer-wise adaptive attention to obtain the sentence representation.

\subsection{Layer-wise Adaptive Attention}

In this work, pre-trained language models are used as the encoder of our model which encodes each token in the input sequence into a fixed-length vector. $h_c^i$ is the representation $[CLS]$ of
layer $i$, and $h_a^i$ is the pooled representation, which is the mean vector of input tokens.

\begin{eqnarray}
\alpha _ i = \frac{W[h_c^i;h_a^i]}{\sum_{j\in N}W[h_c^j;h_a^j]}
\end{eqnarray}
\begin{eqnarray}
H_L = \sum_{i \in N}\alpha _ i [h_c^i;h_a^i]
\end{eqnarray}


The ground truth is $y \in \{0,1 \}$, Thus we re-define the score  $g(P,Q,A_{i})$ as:
\begin{equation}
g(P,Q,A_{i}) = \sigma (WH_L + b)
\end{equation}
This score represents the probability that the model considers the option to be correct. 
Correspondingly, the cross entropy loss function can be re-formulated as:
\begin{eqnarray}
\text{loss} &=& -\sum y \log (g(P,Q,A_{i})) \nonumber \\
&+& (1-y) \log (1-g(P,Q,A_{i})) 
\end{eqnarray}

Finally, to obtain the correct answers to the question, we select the top-n answers with respect to score. Here $n$ denotes the number of correct answers. E.g., $n$=1 in RACE. 

\begin{table*}[ht]
 \caption{Instances of different MRC resources and preprocessed single-choice instances. }
 \centering
 \small
 \resizebox{1\linewidth}{!}{
\begin{tabular}{llll}
\hline
Dataset & Type & Instance & Single-choice Instance \\ \hline
\multirow{3}{*}{Dream} & \multirow{3}{*}{Multi-Choice} & \multirow{3}{*}{\begin{tabular}[c]{@{}l@{}}Context: M: Did you watch TV yesterday\\ evening?\\ F: No, I saw a film instead.\\ Question: What did the man probably do\\ yesterday evening?\\ Choice 1: Watch TV.\\ Choice 2: Saw a film.\\ Choice 3: Read a book.\end{tabular}} & \begin{tabular}[c]{@{}l@{}}M: Did you watch TV yesterday evening?\\ F: No, I saw a film instead. What did\\ the man probably do yesterday evening? \\ Watch TV.\\ label 1\end{tabular} \\ \cline{4-4} 
 &  &  & \begin{tabular}[c]{@{}l@{}}M: Did you watch TV yesterday evening?\\ F: No, I saw a film instead. What did\\ the man probably do yesterday evening?\\ Saw a film.\\ label 0\end{tabular} \\ \cline{4-4} 
 &  &  & \begin{tabular}[c]{@{}l@{}}M: Did you watch TV yesterday evening?\\ F: No, I saw a film instead. What did\\ the man probably do yesterday evening?\\ Saw a film.\\ label 0\end{tabular} \\ \hline
SQuAD & Extractive & \begin{tabular}[c]{@{}l@{}}Context: On January 7, 2012, Beyonce \\ gave birth to her first child, a \\ daughter, Blue Ivy Carter,at Lenox \\ Hill Hospital in New York. ...... her \\ first performances since giving birth \\ to Blue Ivy. \\ Question: Where did Beyonce give birth \\ to her first child? \\ Answer: Lenox Hill Hospital\end{tabular} & \begin{tabular}[c]{@{}l@{}}On January 7, 2012, Beyonce gave birth \\ to her first child, a daughter, Blue \\ Ivy Carter, at Lenox Hill Hospital in \\ New York. ...... her first performances \\ since giving birth to Blue Ivy. Where\\ did Beyonce give birth to her first \\ child? Lenox Hill Hospital.\\ label 1\end{tabular} \\ \hline
\end{tabular}
}
\label{tab:transfer}
\end{table*}

\subsection{Transfer Learning}

In this section, We propose a simple yet effective strategy to transfer knowledge from other QA datasets. As the single-choice model relax the constraints on multi-choice framework, more QA datasets such as SQuAD2.0, ARC, CoQA and DREAM can be used to enhance RACE. It consists of three steps:

(1) we preprocess data with different formats to the same input type as mentioned in section 3. For multiple-choice MRC datasets, like DREAM and ARC, we concatenate each option with corresponding context and question.
And for extractive MRC datasets like SQuAD2.0 and CoQA, we take the context (passage or dialog), one of its questions and corresponding answer as a positive instance for the binary classification, as shown in the Table \ref{tab:transfer}.

(2) We collect and corrupt the preprocessed data from different QA datasets
and then train the binary classification on this mixed data. We find that the model benefits a lot from the large amount of MRC datasets.

(3) Finally, we further finetune the model from step 2 on the raw RACE data to adapt the model parameters to the task.

\begin{table*}[ht]
    \centering
    \setlength{\tabcolsep}{6pt}
    \caption{Details of different MRC resources. ``\#Instance" refers to the number of true training samples built by different resources.}
    \begin{tabular}{lcccccc}
    \hline 
        Models & RACE & DREAM & SQuAD2.0 & CoQA & ARC & Crawl \\ \hline
        \#Article  & 27,933 & 6,444 & 130,319 & 8,399 & - &  - \\
        \#Question  & 97,687 & 10,197 & - & 127,000 & 7787 &  -  \\
        \#Answer per Question  & 4 & 3 & 1 & - & 4 &  -  \\
        \#Word per Article & 321.9 & 85.9 & - & 271 & - &  -  \\
        \#Instance  & 351,464 & 15,470 & 86,835 & 90,000 & 20,784 &  446,095  \\
        \hline
    \end{tabular}
    \label{tab:data}
\end{table*}

\section{Experiments}

\subsection{Dataset}

We use two MMRC datasets as the target datasets: RACE \cite{lai-etal-2017-race} and DREAM \cite{sun-etal-2019-dream}. For MMRC task, the evaluation criteria is accuracy, $acc = N^+ / N $, where $N^+$ denotes the number of correct predictions and the total number of evaluation examples.
In the transfer learning stage, we also consider other MRC tasks. Specifically, we consider SQuAD2.0 \cite{rajpurkar-etal-2016-squad}, ARC \cite{Clark2018ThinkYH}, CoQA \cite{reddy-etal-2019-coqa}.  For all datasets, we use the official train/dev/test splits. We give a brief description of these datasets.

\textbf{RACE} \cite{lai-etal-2017-race} is a dataset collected from middle and high school English exams in China. RACE has a wide variety of question types such as summarization, inference, deduction and context matching. It contains articles from multiple domains (i.e. news, ads, story) and most of the questions need reasoning. 

RACE is created by domain experts to test students' reading comprehension skills, consequently requiring non-trivial reasoning techniques. Each article in RACE has several questions and the questions always have 4 candidate answers, one answer and three distractors.

\textbf{DREAM} is multiple-choice dialogue-based reading comprehension examination dataset. It article is a dialog and each question has only three options. DREAM is a
challenging dataset because 85\% of the questions require reasoning beyond a single sentence, and 34\% of the questions involve commonsense knowledge.

\textbf{SQuAD2.0 and CoQA} are extractive MRC tasks, the articles of which are wiki passages and dialogs. Their questions do not have candidate answers, instead, participants are asked to extract the answer from the passage.

\textbf{ARC} is the largest public-domain multiple-choice dataset that consists of natural and grade-school science question. It is partitioned into a Challenge Set and an Easy set.

Although we have transferred as much data as we can, the MMRC task still suffers data insufficiency problem. Thus we crawl different kind of MRC data from the website. Table \ref{tab:data} lists all the resources we use.

\begin{table}[t]
    \centering
    \caption{Results in accuracy ($\%$) on  test set of RACE dataset.}
    \resizebox{\linewidth}{!}{
    \begin{tabular}{ll}
    \hline 
        Models & Test \\ \hline
        $ \rm XLNet_{xxlarge}$ + DCMN+ &   82.8  \\ 
        RoBERTa \cite{liu2019roberta} &    83.2   \\
        DCMN+ (ensemble) \cite{Zhang_Zhao_Wu_Zhang_Zhou_Zhou_2020}  &   84.1 \\
        ALBERT (single) \cite{Lan2020ALBERT} &    86.5   \\
        RoBERTa + MMM \cite{Jin_Gao_Kao_Chung_Hakkani-tur_2020} & 85.0 \\
        $ \rm T5^{*}$ \cite{2020t5} &  87.1 \\
        UnifiedQA \cite{2020unifiedqa} & 89.4 \\
        ALBERT (ensemble) \cite{Lan2020ALBERT} &   89.4   \\
        ALBERT + DUMA (single) \cite{zhu2021duma} &   88.0   \\
        ALBERT + DUMA (ensemble) \cite{zhu2021duma} &  89.8   \\
        Megatron-BERT (single) \cite{shoeybi2019megatron}   &  \bf 89.5   \\
        Megatron-BERT (ensemble) \cite{shoeybi2019megatron}  &  \bf 90.9   \\
        \hline
        ALBERT baseline        & 87.1   \\
        ALBERT single-choice   & 87.9   \\
        + transfer learning   &  88.3   \\
        + layer-wise adaptive attention  &90.0\\
         + crawled corpus  & \bf 90.7   \\
        Ensemble & \bf 91.4   \\
        \hline
    \end{tabular}
    }
    \label{tab:main-result}
\end{table}

\begin{table}[ht]
\centering
\renewcommand{\arraystretch}{1.0}
\caption{Results in accuracy ($\%$) on test set of DREAM dataset.}
\resizebox{\linewidth}{!}{
\begin{tabular}{lll}
\hline 
Models  & Test \\ \hline
$\rm BERT_{large}$   & 66.9 \\
$\rm BERT_{large}$ + WAE \cite{Kim_Fung_2020}   & 69.0 \\
$\rm XLNet_{large}$   & 72.0 \\
$\rm RoBERTa_{large}$   & 85.0 \\
$\rm RoBERTa_{large}$ + MMM \cite{Jin_Gao_Kao_Chung_Hakkani-tur_2020} &  88.9 \\
$\rm ALBERT_{xxlarge}$   & 88.5 \\
$\rm ALBERT_{xxlarge}$ + Retraining \cite{ju-etal-2021-enhancing} & 90.0 \\
$\rm ALBERT_{xxlarge}$ + DUMA\\ + Retraining  & 90.2 \\
$\rm ALBERT_{xxlarge}$ + DUMA  & 90.4 \\
$\rm ALBERT_{xxlarge}$ + HRCA+ \cite{zhang-yamana-2022-hrca}  & 91.6 \\
$\rm ALBERT_{xxlarge}$ + DUMA\\ + Multi-Task Learning  & 91.8 \\
$\rm ALBERT_{xxlarge}$ + HRCA+\\ + Multi-Task Learning  & \bf 92.6 \\ \hline

ALBERT single-choice  & 90.0 \\
+ layer-wise adaptive attention  &91.8\\
+ transfer learning & \bf 93.2  \\ \hline
\end{tabular}
}
\label{tab:dream-result}
\end{table}

\subsection{Experimental Settings}

Our implementation was based on Transformers\cite{wolf-etal-2020-transformers}\footnote{https://github.com/huggingface/transformers}. We use the ALBERT-xxlarge as encoder. For hyperparameters, we follow \cite{Lan2020ALBERT}, except that we set the learning rate to 1e-5 and the warmup steps to 2000. Because we find this is better for the huggingface ALBERT-xxlarge model. After adding the other resources, we do not use a fixed ``Training Steps", the training steps after two epochs and the warm-up step is 10\% of the total training steps. All the models are trained on 4 NVIDIA V100 GPUs.

\textbf{Baseline} Our baseline is the original huggingface ALBERT-xxlarge model with the default multi-choice strategy. The hyper parameters follow the description above. In addition, we compare our model with many other public results from both papers or the leaderboard. 
The other PrLMs used for comparison include BERT\cite{devlin-etal-2019-bert}, XLNet\cite{NEURIPS2019_dc6a7e65}, RoBERTa\cite{liu2019roberta}, T5\cite{2020t5}, ALBERT\cite{Lan2020ALBERT}, Megatron\cite{shoeybi2019megatron}.
The other methods based on PrLMs include WAE\cite{Kim_Fung_2020}, DCMN\cite{Zhang_Zhao_Wu_Zhang_Zhou_Zhou_2020}, MMM\cite{Jin_Gao_Kao_Chung_Hakkani-tur_2020}, UnifiedQA\cite{2020unifiedqa}, DUMA~\cite{zhu2021duma}, Retraining\cite{ju-etal-2021-enhancing}, HRCA+\cite{zhang-yamana-2022-hrca}.

\textbf{Data pre-processing} Every sentence in the DREAM dataset begins with the speaker's gender, i.e., man and woman. For example, in utterance ``M: Did you watch TV yesterday evening? F: No, I saw a film instead.”, ``m” is the abbreviation of ``man”, ``w" and ``f" represent ``woman". However, the full names such as ``woman” and ``man" are used in the questions. Therefore, We apply data pre-processing by replacing ``w” or ``f” with ``woman” and ``m” with “man” to maintain the consistency of gender representations.

\subsection{Results}
In this study we adopted accuracy as the evaluation metric.
Table \ref{tab:main-result} shows the results on the RACE datasets. The top part of the table lists the results from the current leaderboard \footnote{http://www.qizhexie.com/data/RACE\_leaderboard.html} and papers. Megatron-BERT \cite{shoeybi2019megatron} achieves the best single and ensemble results. It is a variant of BERT\cite{devlin-etal-2019-bert} with 3.9 billion parameters which is almost 40 times bigger than ALBERT-xxlarge, so most methods are based on ALBERT model.

The results of our models are listed below the table. Our ALBERT baseline yields better results than the original ALBERT due to the different choices of hyper-parameters. Compared with the baseline, our single-choice model achieves an improvement of 0.8 in accuracy, which shows that single-choice is better than multi-choice under the ALBERT-xxlarge model. After transferring knowledge from other MRC datasets(excluding crawl corpus), we get another 0.4 more score. With the help of layer-wise adaptive attention, our single model achieves 90\% which surpasses Megatron-BERT \cite{shoeybi2019megatron} and becomes the new state-of-the-art single model results. When adding the web crawl corpus into transfer learning, our single model gets a final score as high as 90.7\%. This illustrates that single-choice model is easy to incorporate other resources and we achieve this by a simple transfer learning strategy. Our ensemble model gets the best score of 91.4\%.

We also evaluate our method on the DREAM dataset. The results are summarized in Table \ref{tab:dream-result}. In the table, we first report the accuracy of the SOTA models in the leaderboard.  Compared with the baseline, our single-choice model achieves 1.5 more score, which shows that single-choice is better than multi-choice. Previous best methods is HRCA+ + Multi-Task Learning, which is HRCA+ 
By applying layer-wise adaptive attention and transfer learning, our model exceeds the Previous best methods HRCA+ + Multi-Task Learning, which is combination of the HRCA+ and Multi-Task Learning \cite{Jin_Gao_Kao_Chung_Hakkani-tur_2020}.
In RACE and DREAM datasets, our model both exceeds the baseline ALBERT model and achieves state-of-the-art performance.

\subsection{Ablation study}

\begin{table}[h]
\centering
\renewcommand{\arraystretch}{1.0}
\caption{Transfer learning ablation study on DREAM dataset.}
\begin{tabular}{llll}
\hline 
Models & Dev & Test & $\downarrow$ \\ \hline
ALBERT single-choice \\
+ transfer learning & 92.7 & 93.2 & \_  \\ \hline
- ARC & 92.7 & 92.8  & 0.4\\
- CoQA & 92.2 & 92.8  & 0.4\\
- SQuAD & 92.1 & 92.4 & 0.8\\
- RACE & 88.8 & 89.8  & 3.4\\
- DREAM & 89.9 & 90.6 & 2.6\\ \hline
\end{tabular}
\label{tab:transfer-result}
\end{table}

\begin{table}[t!]
  \caption{Two examples of incorrect predictions in RACE.}
  \label{tab:err_example}
  \begin{tabular}{p{7cm}}
  \toprule
  \textbf{Passage}:Here are some ideas for learning English well ... Listen to English every day. Listen to English radio ... Practise the conversations. Make up conversations ... You'd better use beginner textbooks. Reading English stories. ... and so on. Try English Club.com for young learners. Write down new words. Start a new word notebook. Write words in (A...B...C). Make some sentences. Try to use an English\-English dictionary. Keep an English diary. Start with one sentence. ... Write another sentence tomorrow. \\
  \textbf{Question}: The writer gave us \_\_\_ ideas for learning English faster. \\
  \textbf{Correct Answer}: Six. \\
  \textbf{Predicted Answer}: Seven. \\
  \midrule
  \textbf{Passage}: As many as 10 of the 17 kinds of penguins may be in danger of disappearing ...  Widespread fishing,exploration for oil and oil leaks also make penguins be in danger... Several years ago,oil leaking from a ship hurt 40\% of the penguins in South Africa. The penguins became covered with oil. But thousands of people helped clean and treat the birds well. Then they returned the penguins to the wild. Now these South African penguins are reproducing in higher numbers than before the oil leaking.	   \\
  \textbf{Question}: Why are the penguins in South Africa reproducing in higher numbers than before the oil leaking?     \\
  \textbf{Correct Answer}: Because people treated the penguins well.   \\
  \textbf{Predicted Answer}: Because people helped the penguins.   \\
  \bottomrule
  \end{tabular}
\end{table}

Transfer learning with other datasets can improve the effect on the target dataset. We also want to know how much each dataset contributes to the target dataset. By removing the specific dataset in training phase, we obtain the results in the Table \ref{tab:transfer-result}. The performance loss caused by different datasets may be related to the size and type of datasets. Removing the RACE dataset causes the greatest performance loss because it is the same type as DREAM, and has the largest number of samples. The performance degradation caused by removing the dream's own training set is 2.6, smaller than RACE because its size is not that large. SQuAD and CoQA are of different types from DREAM, so they have less impact on performance loss. ARC is also an MMRC dataset, but ARC and RACE are from the student exams, which leads to the least impact when it is removed.

\subsection{layer-wise adaptive attention analysis}

Why is it better to use layer-wise adaptive attention than just use [CLS] vector of the last layer? 
As mentioned in the study of interpreting BERT \cite{ramnath-etal-2020-towards}, different layers have different concerns. The initial layers primarily focus on question words that are present in the passage and  learn some form of passage-query interaction before arriving at the answer.
The later layers, reduce the focus on question words, and pay more attention on the supporting words that surround the answer span. Besides, Later layers pay importance to words of the same type as the answer, even though the words are not related or necessary to answer the
question. It means BERT distributes its focus over confusing words in later layers. Therefore, it is not efficient enough to just use the vector of last layer. Layer-wise adaptive attention can capture useful information across all layers.

\subsection{Error analysis}

We show two examples of incorrect predictions in the RACE dataset in the Table \ref{tab:err_example}. The first example requires the model to classify ideas, such as listen, practise, read, write etc. These ideas may have similar semantics, and people will also be confused when counting. In the second example, the incorrect predicted answer could also be argued as correct as the incorrect option is also plausible. The relevant sentence in the article is ``But thousands of people helped
clean and treat the birds wel``. The answers ``treated the penguins
well`` and ``helped the penguins`` are very confusing.

\section{Conclusions}

In this paper, we propose a single-choice model for MMRC that consider the options separately, which gets rid of the multi-choice framework and can transfer knowledge from other MRC dataset.
Experiments results demonstrate that our method achieves significant improvements and by taking advantage of other MRC datasets, we achieve a new state-of-the-art performance.

\bibliography{emnlp-ijcnlp-2019}
\bibliographystyle{acl_natbib}

\end{document}